\documentclass[a4paper,conference]{IEEEtran}
\usepackage{algorithm}  
\usepackage{algorithmic}  
\usepackage{multicol}

\ifCLASSINFOpdf
\else
\fi
\begin{document}
%
\title{Adaptive Optimizer for Automated Hyperparameter Optimization 
	Problem }

\author{\IEEEauthorblockN{Huayuan Sun}
\IEEEauthorblockA{sunhuayuan@stu.pku.edu.cn\\
Peking University\\ Beijing, China
}
}


%


\maketitle

\begin{abstract}
The choices of hyperparameters have critical effects on the performance of machine learning models. In this paper, we present a general framework that is able to construct an adaptive optimizer, which automatically adjust the appropriate algorithm and parameters in the process of optimization. Examining the method of adaptive optimizer, we product an example of using genetic algorithm to construct an adaptive optimizer based on Bayesian Optimizer and 
compared effectiveness with original optimizer.
Especially, It has great advantages in parallel optimization.
\end{abstract}


%
\IEEEpeerreviewmaketitle

\section{Introduction}
The effect of machine learning model largely depends on the selection of super parameters, which has attracted more and more attention in recent years. The problem can be formulated as solving the following minimization problem 
$$
\mathop{\arg\min}\limits_{x\in \Omega} f(x),
$$
where $f$ is a specific metrics function, $\Omega$ is the hyperparameter space and $x$ is the parameter in $\Omega$. 

There are many hyperparameter optimization methods applied in practice. Random Search and Grid Search are  simple and straightforward methods to find optimal parameters hyperparameter. The advantage is that the algorithm is clear and suitable for parallel operation. However, they need to try enough parameter combinations, which is very time-consuming. The other is based on Sequential Model-Based Global Optimization [2-4], such as Bayesian optimization (BO) [1] and Tree-structured Parzen Estimator (TPE), which can use relatively few steps to achieve better results. While, they have poor support for parallel operation. In addition, different optimization models perform great differences in different optimization tasks, which may be related to the machine learning algorithm used, the distribution of training data, the selection of parameter space and evaluation function. In addition, different optimization models are suitable in different optimization stages. For example, we except more exploration in the early stage of optimization, and later we need more local search, which is similar to the decreasing learning rate in machine learning. The same is true for different optimizer.

In this paper, we propose a general framework, which can construct an adaptive optimizer and automatically adjust the appropriate algorithms and parameters in the optimization process. Our approach allow to automatically adjust the appropriate algorithm and parameters in the optimization process, and it has good support for parallel operation. In short, we introduce the variable $\phi$ into the optimization objective, and change original optimization problem into the form
$$
\mathop{\arg\min}F(\phi,N) = \mathop{\arg\min}\limits_{1\le n \le N} f(\phi(\theta_n,n)),
$$
where  $\phi:(\theta_n,n)\to x \in \Omega$ represents the optimization function which suggest parameter $x$ in round $n$. $N$ is the maximum number of iterations. Thus, the core problem is to determine the parameters $\theta_n$. Among other algorithms, $\phi$ and $\theta$ is a definite function and parameter.

To show the benefit of our adaptive optimizer compared to fixed optimizer in the process of optimization, we consider a parameter optimization for lightgbm by simple Bayesian optimizers and our approach which is adaptive Bayesian optimizers respectively.
Bayesian optimizationis a classical and effective method to solve the above problems, for the derivative of neural network hyperparameters cannot be obtained and Bayesian optimization is gradient-free global optimization. It contains two core components,  probabilistic surrogate model and acquisition function. The probabilistic surrogate model is to approximatly estimate $f(x)$ based on historical evaluation, commonly including 
Gaussian Processes (GP), Eextra Trees (ET), Random Forests (RF), Gradient Boosted Regression Trees (GBRT). 

And acquisition function measures the utility by trading off exploration and exploitation, such as \\
(1) Lower Confidence Bound (LCB):
$$
x_{t} = \mathop{\arg\min} \mu_{t-1} - \beta^{1/2} \sigma_{t-1}
$$
(2) Expected Improvement (EI):
$$
x_t = \mathop{\arg\min} E_{f(x) \sim N(\mu_{t-1},\sigma_{t-1}^2)} \left[ \min(f(x)-f_{t-1},0)\right]
$$
(3) Probability of Improvement (PI):
$$
x_t = \mathop{\arg\min} P(f(x)\leq f_{t}(x))
$$

In the following sections, we first give
a brief introduction to our  algorithm framework in section two. Afterwards, we state our adaptive approach for hyperparameters optimization problem by genetic algorithm on two examples$\footnote{https://github.com/shyjin/Example-of-GA-for-Adaptive-Optimizer }$.

\begin{algorithm*}  
	\caption{Adaptive Optimizer}  
	\label{alg:A}  
	\begin{algorithmic}  
		\STATE 1.  Generate initial points\\
		2. Calculate rewards $f_{j}^*$ of each optimizer $j$ in the set of meta optimizers $\{\phi_j(\theta_n,n)\}$\\
		3. Choose different $N_{s}$ optimizers based on $f^*$ \\
		4. Suggest next point $x_i$ according to the optimizer  $\phi(\theta_i,step_i)$, and get the value $f(x_i)$\\
		5. Repeat step 2 to 4 until termination criterion is reached\\ 
		6. Output the best record

	\end{algorithmic}  
\end{algorithm*}  

\section{Adaptive Optimizer}

In this section, we introduce our approach of solving a hyperparameters optimization problem. For convenience, we will not distinguish between algorithms and parameters in the future. As long as the algorithm or parameters are not exactly the same, we regard them as different optimizers.

Our method is basically the same as the usual algorithm, except that the step 2 and 3 of adjusting the optimizer is added. We take the adjusted $f_n$ returned by the optimizer in step $n$ as a reward to optimize the optimizer. Then, there is a problem that the effectiveness of the reward function is different at different iterations. For example, in the 15th round, we use optimizer A to get a set of parameters with model accuracy of 0.9, which is completely different from that in the 90th round with optimizer B. In this example, optimizer A is obviously better than optimizer B for optimizer B gets more information, but the accuracy is not improved. Then, if optimizer B achieves an accuracy of 0.92 in the 90th round, how to make a consistent comparison between the two optimizers is the focus of our discussion.

We defined adjusted reward function as 
$$
f^* = \max\left(\epsilon,\frac{f(x) - \frac 1 n \sum f_k} {std(f_1,f_2,...,f_n)}  - \alpha n\right),
$$
where $\epsilon > 0 $ is the minimum weight , $n$ is the  number of current iteration and $alpha=\frac {c} {N}$, $c$ is an non-negative constant like 1.96 or 2.23 and $N$ is the maximum number of iterations. In fact, any non-decreasing function relative to $n$ is available. Considering that the marginal benefit of previous iterations is great, non-convex functions are more suitable than linear function. 

Suppose the number of suggestion $N_s=1$, which means that we try once on objective function and get rewards, then we update model. The simplest way to select the next round of optimizer is to select it on a probabilistic basis. For any fixed optimizer, its probability of selection is based on the largest weight it has ever received. If it is not selected, we can set the average weight of the population. Sum the selection weights of all the optimizers and calculate the corresponding proportion, which is the probability of selection. We give an example in Section 3.

For more complex problems, we ask more than one set of parameters and sent to objective function. Then, running the $N_s>1$ set of parameters optimization tasks in $N_s$ multithreads. Intelligent optimization algorithms such as genetic algorithm can be used. The core approach is to use the adjusted reward function $f^*$ above as fitness function, and cross-over and mutation are added when selecting the optimizer for the next iteration.

\section{Examples}
In order to test the effect of our method, we choose to build two practical hyperparametric optimization problems for testing. General test and parallel test are carried out respectively. We choose the most downloaded dataset "credit-g" on OpenML. This dataset classifies people described by a set of attributes as good or bad credit risks. And we divide the train dataset and test dataset according to the proportion of 70\% and 30\%. The prediction model is lightgbm, and AUC of the model in the test set is taken as the evaluation function. AUC is a commonly used evaluation metrics in classification problems. Five parameters and ranges commonly used in lightgbm algorithm are selected as the hyperparameter space. The hyperparameters are 
\begin{enumerate}
	\item num\_leaves: Maximum tree leaves for base learners. The space is [4, 100], int type.
	\item min\_child\_samples: Minimum number of data needed in a child. The space is [1, 100], int type. 
	\item n\_estimators: Number of boosted trees to fit. The space is [1, 100], int type. 
	\item subsample: Subsample ratio of the training instance. The space is [0.1, 1], float type. 
	\item  colsample\_bytree: Subsample ratio of columns when constructing each tree. The space is [0.1, 1], float type.
\end{enumerate}

The maximum number of iterations for each optimization is $N$, and the number of test parameters requested for each iteration is $N_s$. We take the best score achieved in all iterations as the final score of the model. Without losing generality, as we return the score to the optimizer, we will use $-f^*$ and change the problem into solving the minimum value. In addition, we repeated the experiment 10 times, and selected different random seeds each time.

\subsection{Example:  Simple Task, $N_s=1$}
In this experiment, we set the maximum number of iterations to $N=100$ and return $N_s=1$ set of parameter verification results each time. Our base optimizers are  Bayesian optimizers with GP, RF, ET and GBRT surrogate models respectively. In each round of iteration, we will allocate the weight of the next round of selection according to the historical scores of the four base optimizers, and then install a certain proportion to select one of them. The specific steps have been described in section two.

We use four base optimizers to perform optimization tasks separately, and other settings are the same. The results are as follows: the average AUC of the final 10 experiments of  GP, RF, ET and GBRT are 0.7613, 0.7591,0.7638 and 0.7505 respectively; The average AUC of our algorithm is 0.7649. Our algorithm is better than the four base classifiers. It should be emphasized that although we have only increased by 0.0011 compared with the best performing base optimizer ET, our advantage is that we don't need to deliberately select the optimizer. We only need to put a group of optimizers into the model to achieve the effect no less than any optimizer. In reality, the optimization problem is very complex. It is difficult for us to know which optimizers are suitable for this problem before the actual test. Even at different stages of optimization, the performance effects of different optimizers are different. The proposed algorithm can automatically adjust the optimizer in the iterative process

\begin{table}[h]
\setlength{\tabcolsep}{3mm}{
	\begin{tabular}{|l|l|l|l|l|l|}
		\hline
		& \textbf{GP} & \textbf{RF} & \textbf{ET} & \textbf{GBRT} & \textbf{MyAlgo} \\ \hline
		\textbf{seed=0}    & 0.7372      & 0.7777      & 0.7758      & 0.7887        & 0.7727          \\ \hline
		\textbf{seed=1}    & 0.7552      & 0.7679      & 0.7793      & 0.7887        & 0.7812          \\ \hline
		\textbf{seed=2}    & 0.757       & 0.7815      & 0.7900      & 0.7797        & 0.7614          \\ \hline
		\textbf{seed=3}    & 0.778       & 0.7782      & 0.7927      & 0.7674        & 0.7676          \\ \hline
		\textbf{seed=4}    & 0.7805      & 0.7687      & 0.7668      & 0.7865        & 0.7129          \\ \hline
		\textbf{seed=5}    & 0.7786      & 0.7394      & 0.7917      & 0.7339        & 0.781           \\ \hline
		\textbf{seed=6}    & 0.7671      & 0.6817      & 0.7168      & 0.6878        & 0.7891          \\ \hline
		\textbf{seed=7}    & 0.7404      & 0.7684      & 0.7266      & 0.7887        & 0.7371          \\ \hline
		\textbf{seed=8}    & 0.7715      & 0.7465      & 0.7611      & 0.6465        & 0.7554          \\ \hline
		\textbf{seed=9}    & 0.7473      & 0.7809      & 0.7366      & 0.7365        & 0.7907          \\ \hline
		\textbf{auc\_mean} & 0.7613      & 0.7591      & 0.7638      & 0.7505        & 0.7649          \\ \hline
	\end{tabular}
}
\caption{In this table, the algorithms to be compared with are the base Optimizer with  GP, RF, ET and GBRT surrogate model. }
\end{table}

\subsection{Example:  Parallel Tasks, $N_s>1$}
In this experiment, we set the maximum number of iterations to $N=50$ and take $N_s=3$ set of parameter verification results each time. Our base optimizers are also  Bayesian optimizers with GP, RF, ET and GBRT surrogate models respectively, and the difference is we use four different acquisition function as well. They are LCB, EI, PI and GP Hedge (probabilistically choose one of the above three acquisition functions at every iteration), and we have 16 base optimizers for crossing surrogate models and acquisition function. Another difference is the method to use to sample multiple points. the constant liar strategy is used after suggesting one point. With this strategy a copy of optimizer is created, which is then asked for a point, and the point is told to the copy of optimizer with some fake objective, the next point is asked from copy, it is also told to the copy with fake objective and so on. The type of lie defines different flavours of strategies. We use minimum fake objective here, which means set fake objective as the minimum value.

Here, we use genetic algorithm as the selection algorithm of the next round of iterative optimizer. Other parts are the same as before and will not be repeated. We calculate fitness according to $f^*$. Here for base optimizer $j$, we make a slight adjustment and use 
$$
f^*_j = \max\left(\epsilon,\frac{f_j - \frac 1 n \sum f_k} {std(f_1,f_2,...,f_{n*N_s})}  - \alpha n - b\right),
$$
 as fitness, where $b$ is a positive constant and we prefer to focus on the optimizer which have good fitness if $b$ is larger. The weight of each base optimizer is the larger of its corresponding fitness ratio and a fixed value. $k$ is the number of choosing optimizer $j$.  In particular, we only select the historical maximum as the result of one base optimizer, and then select four optimizers as parents according to the corresponding weight. After retention, crossover and mutation, take the first three as the next round optimizer.

\begin{algorithm}  
	\caption{Adaptive Optimizer with Genetic Algorithm  }  
	\label{alg:B}  
	\begin{algorithmic}  

		\STATE 1.  Generate initial points\\
		2. Calculate rewards $f_{j}^*$ of each optimizer $j$ in the set of meta optimizers $\{\phi_j(\theta_n,n)\}$\\
		3. Generate  $N_s$ Optimizers for next iteration\\
    \  3.1 Choose different $NG$ optimizers based on $f^*$ as parents \\
	\ 	3.2 Retention with probability $\epsilon_r$ \\
	\ 	3.3 Crossover surrogate model and acquisition function \\
	\   3.4 Mutate with probability $\epsilon_m$ \\
	\   3.5 Check and replace the duplicate child  \\ 
		4. Suggest next point $x_i$ according to the optimizer  $\phi(\theta_i,step_i)$, and get the value $f(x_i)$\\
		5. Repeat step 2 to 4 until termination criterion is reached\\ 
		6. Output the best record

	\end{algorithmic} 

\end{algorithm}  

To compare our algorithm, we test the 16 base optimizers respectively as well, combination of four surrogate models and four acquisition functions. We calculate some indicators to measure the results, including the mean and median of AUC for the effect of the optimizer, and the variance of AUC for the stability of the optimizer. In Table 2, the results are shown in order of mean value from large to small. The mean and median results of our algorithm are ranked 3rd, and the variance is ranked 4th. The minimum value can be used to measure the effect of the optimizer in the worst case, ranking 3rd place among all optimizers.

Generally speaking, compared with all base optimizers, our algorithm is above average in optimization effect and stability, and can easily run parallel tasks, that is,  $N_s$ tasks executed in each round of iteration. For the base classifier we use, these $N_s$ prediction points must be executed in turn during each round of optimization. Therefore, our algorithm has certain advantages in parallel efficiency. The effect of the model is affected in the following aspects:
\begin{table}[]
	\setlength{\tabcolsep}{1mm}{
	\begin{tabular}{|c|c|c|c|c|c|}
		\hline
		\textbf{Optimizer} & \textbf{auc\_mean} & \textbf{auc\_std} & \textbf{auc\_max} & \textbf{auc\_min} & \textbf{auc\_median} \\ \hline
		RF\_LCB         & 0.7918 & 0.0076 & 0.8055 & 0.7789 & 0.7904 \\ \hline
		ET\_PI          & 0.7895 & 0.012  & 0.8014 & 0.7582 & 0.7921 \\ \hline
		myalgo\_nan     & 0.7891 & 0.0069 & 0.7973 & 0.7772 & 0.7911 \\ \hline
		ET\_LCB         & 0.7886 & 0.0137 & 0.8003 & 0.7508 & 0.7915 \\ \hline
		GP\_LCB         & 0.7862 & 0.0049 & 0.7919 & 0.7748 & 0.7863 \\ \hline
		RF\_PI          & 0.7855 & 0.0084 & 0.7968 & 0.7737 & 0.7859 \\ \hline
		GP\_gp\_hedge   & 0.7852 & 0.0052 & 0.7944 & 0.7787 & 0.7841 \\ \hline
		RF\_gp\_hedge   & 0.785  & 0.0128 & 0.8038 & 0.7638 & 0.7872 \\ \hline
		ET\_gp\_hedge   & 0.7842 & 0.0199 & 0.8038 & 0.7429 & 0.7928 \\ \hline
		GP\_EI          & 0.7821 & 0.0074 & 0.7927 & 0.7695 & 0.7849 \\ \hline
		GP\_PI          & 0.7814 & 0.0045 & 0.7878 & 0.7752 & 0.7813 \\ \hline
		GBRT\_LCB       & 0.7787 & 0.0212 & 0.7978 & 0.7292 & 0.7862 \\ \hline
		GBRT\_gp\_hedge & 0.7743 & 0.0326 & 0.8012 & 0.7059 & 0.7934 \\ \hline
		RF\_EI          & 0.7648 & 0.0661 & 0.7924 & 0.5778 & 0.788  \\ \hline
		GBRT\_PI        & 0.7631 & 0.053  & 0.8055 & 0.6219 & 0.7817 \\ \hline
		ET\_EI          & 0.7572 & 0.0394 & 0.791  & 0.6735 & 0.7658 \\ \hline
		GBRT\_EI        & 0.7474 & 0.0338 & 0.8055 & 0.7    & 0.7503 \\ \hline
	\end{tabular}
}
\caption{In this table, the algorithms to be compared with combination of four surrogate models and four acquisition functions } 
\end{table}
\begin{enumerate}
	\item Differences of base optimizers: similar base optimizers will generate similar recommended parameters in each iteration, which will greatly affect the effect of the model. Adding different kinds of optimizers will greatly improve.\\
	\item Maximum number of iterations N: The selection of optimizer needs the support of historical data. If N is relatively small, due to the insufficient amount of data, it will be greatly affected by random shake. \\
	\item $N_s$: As mentioned in the first point, when $N_s$ is too large, it is easy to make the selected point close in the same round. Enough different base optimizers need to be added.\\

\end{enumerate}
\section{Conclusion}
With this paper, we have introduced a new methodology of automated hyperparameter optimization. Machine learning algorithms are generally time-consuming, so it is difficult to test which algorithms are more suitable for hyperparametric optimization tasks. By using our method, we can automatically adjust the optimizer to adapt to the task in the optimization process. Moreover, a series of algorithms that do not support parallel tasks can be combined to establish parallel tasks. Using a single optimizer is prone to particularly poor performance, and multiple validation takes time much take. The method is more suitable for difficult and time-consuming optimization tasks.




\end{document}